\documentclass{article} % For LaTeX2e
\usepackage{iclr2024_conference,times}

\usepackage{amsmath,amsfonts,bm}

\def\eqref#1{equation~\ref{#1}}
\def\1{\bm{1}}

\def\vtheta{{\bm{\theta}}}

\def\vc{{\bm{c}}}

\def\vf{{\bm{f}}}

\def\vx{{\bm{x}}}

\def\vz{{\bm{z}}}

\DeclareMathAlphabet{\mathsfit}{\encodingdefault}{\sfdefault}{m}{sl}
\SetMathAlphabet{\mathsfit}{bold}{\encodingdefault}{\sfdefault}{bx}{n}

\usepackage{times}
\usepackage{epsfig}
\usepackage{graphicx}
\usepackage{amsmath}
\usepackage{amssymb}
\usepackage{pifont}
\usepackage{booktabs}
\usepackage{arydshln}
\usepackage{color}
\usepackage{colortbl}
\usepackage{subcaption}
\usepackage{multirow}
\usepackage{float}
\usepackage{rotfloat}
\usepackage{capt-of}
\usepackage{longtable}
\usepackage{diagbox}
\usepackage{makecell}
\usepackage{pgfplots}
\usepackage{xspace}
\usepackage{wrapfig}
\usepackage{enumitem}
\usepackage{algorithm}
\usepackage{algorithmic}

\pgfplotsset{compat=1.18}

\definecolor{linkc}{rgb}{0, 0.44, 0.74}
\definecolor{eqc}{rgb}{1, 0, 0}
\usepackage[pagebackref=false,breaklinks=true,colorlinks=True,urlcolor=eqc,citecolor=linkc,linkcolor=eqc,bookmarks=false]{hyperref}

\makeatletter
\DeclareRobustCommand\onedot{\futurelet\@let@token\@onedot}
\def\@onedot{\ifx\@let@token.\else.\null\fi\xspace}

\makeatother

\newcommand{\blfootnote}[1]{\begingroup
\renewcommand\thefootnote{}\footnote{#1}\addtocounter{footnote}{-1}
\endgroup}

\newcommand{\model}{\textsc{PixArt-$\delta$}\xspace}
\newcommand{\pixarta}{\textsc{PixArt-$\alpha$}\xspace}

\definecolor{revise}{HTML}{0071BC}

\definecolor{mycolor_blue}{HTML}{E7EFFA}
\definecolor{mycolor_green}{HTML}{E6F8E0}
\definecolor{mycolor_gray}{HTML}{ECECEC}
\definecolor{pearDark}{HTML}{2980B9}

\title{\model: Fast and Controllable Image \\ Generation with Latent Consistency Models}

\author{%
  Junsong Chen$^{1,2,4}$,\space\space\space
  Yue Wu$^{1}$,\space\space\space
  Simian Luo$^{3}$,\space\space\space
  Enze Xie$^{1\dagger}$,\space\space\space
  \\
  \vspace{0.15cm}
  \textbf{Sayak Paul$^{5}$,\space\space\space
  Ping Luo$^{4}$,\space\space\space
  Hang Zhao$^{3}$,\space\space\space
  Zhenguo Li$^{1}$\space\space\space
  }
  \\
  $^{1}$\normalfont{Huawei Noah's Ark Lab}\quad
  $^{2}$\normalfont{Dalian University of Technology}\quad
  \\
  $^{3}$\normalfont{IIIS, Tsinghua University}\quad
  $^{4}$\normalfont{The University of Hong Kong}\quad
  $^{5}$\normalfont{Hugging Face}\quad
  \vspace{0.15cm}\\
  \tt\small 
  {jschen@mail.dlut.edu.cn}, \tt\small luosm22@mails.tsinghua.edu.cn, \\
  \tt\small{\{wuyue119,xie.enze,Li.Zhenguo\}@huawei.com}
  \vspace{0.15cm}\\
  {\centering Homepage: \url{https://pixart-alpha.github.io/}}
  \\
  {\centering Code: \url{https://github.com/PixArt-alpha/PixArt-alpha}}
  \\
  {\centering Demo: \url{https://huggingface.co/spaces/PixArt-alpha/PixArt-LCM}}
}

\iclrfinalcopy % Uncomment for camera-ready version, but NOT for submission.
\begin{document}
\maketitle

\blfootnote{$\dagger$ Project lead.}

% ****************** final version **********************
\begin{abstract}
This technical report introduces \model, a text-to-image synthesis framework that integrates the Latent Consistency Model (LCM) and ControlNet into the advanced \pixarta model. \pixarta is recognized for its ability to generate high-quality images of 1024px resolution through a remarkably efficient training process. The integration of LCM in \model significantly accelerates the inference speed, enabling the production of high-quality images in just 2-4 steps. Notably, \model achieves a breakthrough 0.5 seconds for generating 1024 $\times$ 1024 pixel images, marking a 7$\times$ improvement over the \pixarta. Additionally, \model is designed to be efficiently trainable on 32GB V100 GPUs within a single day. With its 8-bit inference capability~\citep{diffusers}, \model can synthesize 1024px images within 8GB GPU memory constraints, greatly enhancing its usability and accessibility.
Furthermore, incorporating a ControlNet-like module enables fine-grained control over text-to-image diffusion models. We introduce a novel ControlNet-Transformer architecture, specifically tailored for Transformers, achieving explicit controllability alongside high-quality image generation. As a state-of-the-art, open-source image generation model, \model offers a promising alternative to the Stable Diffusion family of models, contributing significantly to text-to-image synthesis.

\end{abstract}
\section{Introduction}
In this technical report, we propose \model, which incorporates LCM~\citep{luo2023latent} and ControlNet~\citep{zhang2023controlnet} into \pixarta~\citep{chen2023pixart}. Notably, \pixarta is an advanced high-quality 1024px diffusion transformer text-to-image synthesis model, developed by our team, known for its superior image generation quality achieved through an exceptionally efficient training process.

We incorporate LCM into the \model to accelerate the inference. LCM~\citep{luo2023latent} enables high-quality and fast inference with only 2$\sim$4 steps on pre-trained LDMs by viewing the reverse diffusion process as solving an augmented probability flow ODE (PF-ODE), which enables \model to generate samples within ($\sim$4) steps while preserving high-quality generations. As a result, \model takes 0.5 seconds per 1024 $\times$ 1024 image on an A100 GPU, improving the inference speed by 7$\times$ compared to \pixarta. We also support LCM-LoRA~\citep{luo2023lcm} for a better user experience and convenience.

In addition, we incorporate a ControlNet-like module into the \model. ControlNet~\citep{zhang2023controlnet} demonstrates superior control over text-to-image diffusion models' outputs under various conditions.
However, it's important to note that the model architecture of ControlNet is intricately designed for UNet-based diffusion models, and we observe that a direct replication of it into a Transformer model proves less effective. Consequently, we propose a novel ControlNet-Transformer architecture customized for the Transformer model. Our ControlNet-Transformer achieves explicit controllability and obtains high-quality image generation.
\section{Background}

\subsection{Consistency Model}
Consistency Model~(CM) and Latent Consistency Model~(LCM) have made significant advancements in the field of generative model acceleration. CM, introduced by \cite{song2023consistency} has demonstrated its potential to enable faster sampling while maintaining the quality of generated images on ImageNet dataset~\citep{deng2009imagenet}. A key ingredient of CM is trying to maintain the self-consistency property during training (consistency mapping technique), which allows for the mapping of any data point on a Probability Flow Ordinary Differential Equation (PF-ODE) trajectory back to its origin. 

LCM, proposed by \cite{luo2023latent}, extends the success of CM to the current most challenging and popular LDMs, Stable Diffusion~\citep{rombach2022high} and SD-XL~\citep{podell2023sdxl} on Text-to-Image generative task. LCM accelerates the reverse sampling process by directly predicting the solution of the augmented PF-ODE in latent space. LCM combines several effective techniques (e.g, One-stage guided distillation, Skipping-step technique) to achieve remarkable rapid inference speed on Stable Diffusion models and fast training convergence. LCM-LoRA~\citep{luo2023lcm}, training LCM with the LoRA method~\citep{hu2021lora}, demonstrates strong generalization, establishing it as a universal Stable Diffusion acceleration module. In summary, CM and LCM have revolutionized generative modeling by introducing faster sampling techniques while preserving the quality of generated outputs, paving the way for real-time generation applications.

\subsection{ControlNet}

ControlNet~\citep{zhang2023controlnet} demonstrates superior control over text-to-image diffusion models' outputs under various conditions (e.g., canny edge, open-pose, sketch). It introduces a special structure, a trainable copy of UNet, that allows for the manipulation of input conditions, enabling control over the overall layout of the generated image. During training, ControlNet freezes the origin text-to-image diffusion model and only optimizes the trainable copy. It integrates the outputs of each layer of this copy by skip-connections into the original UNet using ``zero convolution" layers to avoid harmful noise interference. 

This innovative approach effectively prevents overfitting while preserving the quality of the pre-trained UNet models, initially trained on an extensive dataset comprising billions of images. 
ControlNet opens up possibilities for a wide range of conditioning controls, such as edges, depth, segmentation, and human pose, and facilitates many applications in controlling image diffusion models.

\section{LCM in \model}
In this section, we employ Latent Consistency Distillation~(LCD) \citep{luo2023latent} to train \model{} on 120K internal image-text pairs. In Sec.~\ref{sec:algorithm_and_modification}, we first provide a detailed training algorithm and ablation study on specific modifications. In Sec.~\ref{sec:train_and_inference}, we illustrate the training efficiency and the speedup of LCM of \model{}. Lastly, in Sec.~\ref{sec:training_details}, we present the training details of \model{}.

\subsection{Algorithm and modification \label{sec:algorithm_and_modification}}

\begin{figure}[!ht]
\centering
\vspace{-1em}
\footnotesize
\includegraphics[width=0.95\linewidth]{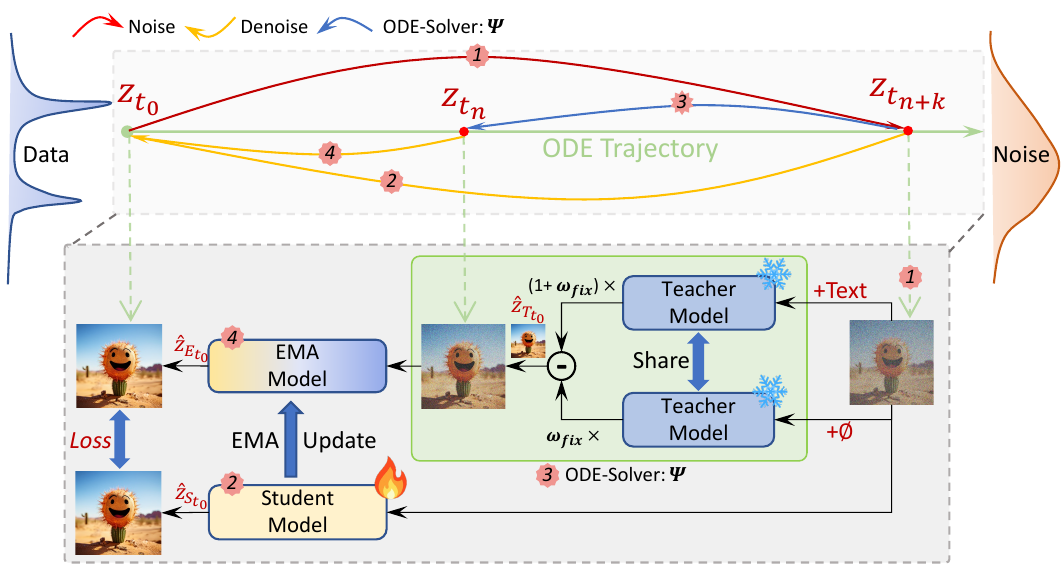}
\caption{Training pipeline of \model{}. The upper section of the diagram offers a high-level overview of the training process, depicting the sequential stages of noise sampling and denoising along a specific ODE trajectory. Sequence numbers are marked on the mapping lines to clearly indicate the order of these steps. The lower section delves into the intricate roles of the pre-trained (teacher) model and the student model, revealing their respective functions within the upper block's training process, with corresponding sequence numbers also marked for easy cross-referencing.}
\label{fig:pixart_lcm_pipeline}
\end{figure}

\noindent \textbf{LCD Algorithm.} Deriving from the original Consistency Distillation~(CD)~\citep{song2023consistency} and LCD~\citep{luo2023latent} algorithm, we present the pseudo-code for \textbf{\model{}} with classifier-free guidance~(CFG) in Algorithm~\ref{alg:pixart-lcm}. Specifically, as illustrated in the training pipeline shown in Fig.~\ref{fig:pixart_lcm_pipeline}, three models – Teacher, Student, and  EMA Model – function as denoisers for the ODE solver $\Psi(\cdot,\cdot,\cdot, \cdot)$, $\vf_\vtheta$, and $\vf_{\vtheta^-}$, respectively. During the training process, we begin by sampling noise at timestep $t_{n+k}$, where the  Teacher Model is used for denoising to obtain $\hat{z}_{T_{t_0}}$. We then utilize a ODE solver $\Psi(\cdot,\cdot,\cdot, \cdot)$ to calculate  $\hat{z}^{\Psi, \omega}_{t_n}$ from $z_{t_{n+k}}$ and $\hat{z}_{T_{t_0}}$. EMA Model is then applied for further denoising, resulting in $\hat{z}_{E_{t_0}}$. In parallel, the Student Model denoises the sample $z_{t_{n+k}}$ at $t_{n+k}$ to derive $\hat{z}_{S_{t_0}}$. The final step involves minimizing the distance between $\hat{z}_{S_{t_0}}$ and $\hat{z}_{E_{t_0}}$, also known as optimizing the consistency distillation objective. 
% 
% Notably, within our pipeline, we substitute the original variable guidance scale $\omega$, which is selected from a designated range [$\omega_{min}$, $\omega_{max}$] in LCM, with a fixed value $\omega_{fix}$ for convenience and better performance.

Different from the original LCM, which selects variable guidance scale $\omega$ from a designated range [$\omega_{min}$, $\omega_{max}$], in our implementation, we set the guidance scale as a constant $\omega_{fix}$, removing the guidance scale embedding operation in LCM \citep{luo2023latent} for convenience.

\begin{algorithm}[H]
    \begin{minipage}{\linewidth}
    \begin{footnotesize}
        \caption{PixArt - Latent Consistency Distillation (LCD)}\label{alg:pixart-lcm}
        \begin{algorithmic}
            \STATE \textbf{Input:} dataset $\mathcal{D}$, initial model parameter $\vtheta$, learning rate $\eta$, {ODE solver $\Psi(\cdot,\cdot,\cdot, \cdot)$}, distance metric $d(\cdot,\cdot)$, EMA rate $\mu$, {noise schedule $\alpha(t),\sigma(t)$, guidance scale $\textcolor{blue}{\omega_{fix}}$, skipping interval k, and encoder $E(\cdot)$}
            \STATE  {Encoding training data into latent space: $\mathcal{D}_z=\{(\vz,\vc)|\vz=E(\vx),(\vx,\vc)\in\mathcal{D}\}$}
            \STATE $\vtheta^-\leftarrow\vtheta$
            \REPEAT
            \STATE Sample $(\vz,\vc) \sim \mathcal{D}_z$, {$n\sim \mathcal{U}[1,N-k]$}
            \STATE Sample $\vz_{t_{n+k}}\sim \mathcal{N}(\alpha(t_{n+k})\vz;\sigma^2(t_{n+k})\mathbf{I})$
            \STATE $\begin{aligned} {\hat{\vz}^{\Psi, \textcolor{blue}{\omega_{fix}}}_{t_n}\leftarrow \vz_{t_{n+k}}+(1+\textcolor{blue}{\omega_{fix}})\Psi(\vz_{t_{n+k}},t_{n+k},t_n,\vc)-\textcolor{blue}{\omega_{fix}}\Psi(\vz_{t_{n+k}},t_{n+k},t_n,\varnothing)}\end{aligned}$
            \STATE $\begin{aligned}\mathcal{L}(\vtheta,\vtheta^-; \Psi)\leftarrow  {d(\vf_\vtheta(\vz_{t_{n+k}},\textcolor{blue}{\textcolor{blue}{\omega_{fix}}},\vc, t_{n+k}),\vf_{\vtheta^-}(\hat{\vz}^{\Psi,\textcolor{blue}{\omega_{fix}}}_{t_n},\textcolor{blue}{\omega_{fix}},\vc, t_n))}\end{aligned}$
            \STATE $\vtheta\leftarrow\vtheta-\eta\nabla_\vtheta\mathcal{L}(\vtheta,\vtheta^-)$
            \STATE $\vtheta^-\leftarrow \text{stopgrad}(\mu\vtheta^-+(1-\mu)\vtheta)$
            \UNTIL convergence
        \end{algorithmic} 
        \end{footnotesize}
    \end{minipage}
\end{algorithm}

\begin{figure}[!ht]
\centering
\vspace{-1em}
\footnotesize
\includegraphics[width=0.95\linewidth]{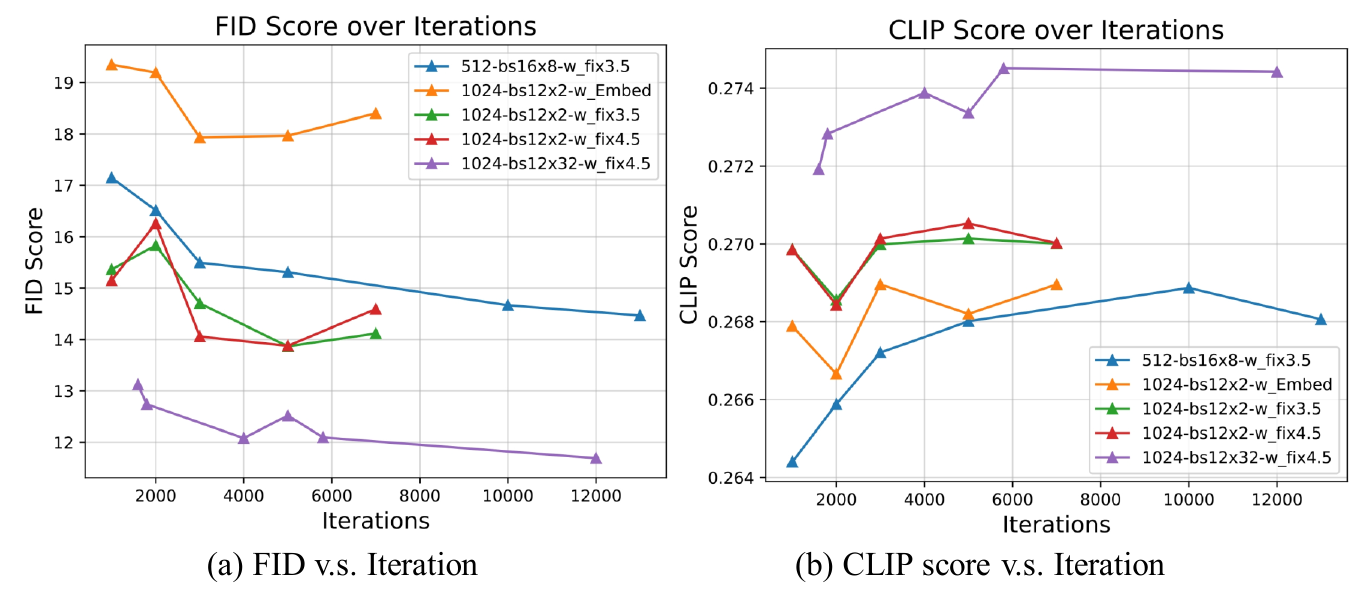}
\caption{Ablation study of FID and CLIP Score on various strategies for classifier-free guidance scale ($\omega$) and their impact on distillation convergence during training.}
\label{fig:fid_clip}
\end{figure}

\begin{figure}[!ht]
\centering
\vspace{-1em}
\footnotesize
\includegraphics[width=0.95\linewidth]{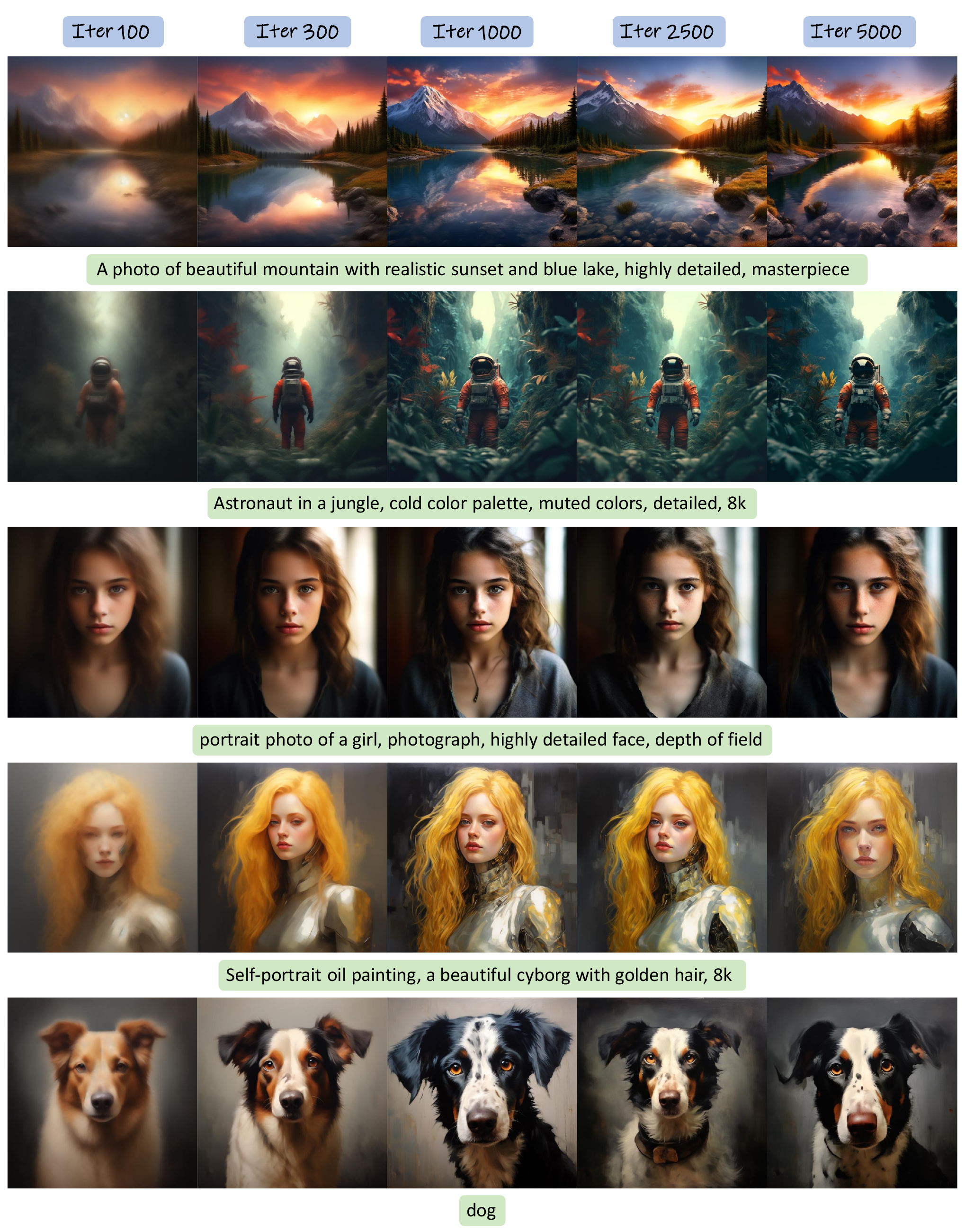}
\caption{The 4-step inference samples generated by \model{} demonstrate fast convergence in LCD training on 2 V100 GPUs with a total batch size of 24. Remarkably, the complete fine-tuning process requires less than 24GB of GPU memory, making it feasible on most contemporary consumer-grade GPUs.}
\label{fig:lcm_converge_samples}
\end{figure}

\noindent \textbf{Effect of Hyper-parameters.} Our study complements two key aspects of the LCM training process, CFG scale and batch size. These factors are evaluated using FID and CLIP scores as performance benchmarks. The terms `$bs$', `$\omega\_{fix}$', and `$\omega\_{Embed}$' in the Fig.~\ref{fig:fid_clip} represent training batch size, fixed guidance scale, and embedded guidance scale, respectively. 
\begin{itemize}
\item
\textbf{CFG Scale Analysis:} Referencing Fig.~\ref{fig:fid_clip}, we examine three distinct CFG scales: (1)~3.5, utilized in our ablation study; (2)~4.5, which yieldes optimal results in \pixarta{}; and (3)~a varied range of CFG scale embeddings ($\omega\_{Embed}$), the standard approach in LCM. Our research reveals that employing a constant guidance scale, instead of the more complex CFG embeddings improves performance in \model{} and simplifies the implementation. 

\item
\textbf{Batch Size Examination:} The impact of batch size on model performance is assessed using two configurations: 2 V100 GPUs and 32 V100 GPUs; each GPU loads 12 images. 
As illustrated in Fig.~\ref{fig:fid_clip}, our results indicate that larger batch size positively influences FID and CLIP scores. However, as shown in Fig.~\ref{fig:lcm_converge_samples}, \model{} can also converge fast and get comparable image quality with smaller batch sizes.

\item
 \textbf{Convergence:} Finally, we observe that the training process tends to reach convergence after approximately 5,000 iterations. Beyond this phase, further improvements are minimal.
\end{itemize}

\begin{figure}[!ht]
\centering
\vspace{-1em}
\footnotesize
\includegraphics[width=0.98\linewidth]{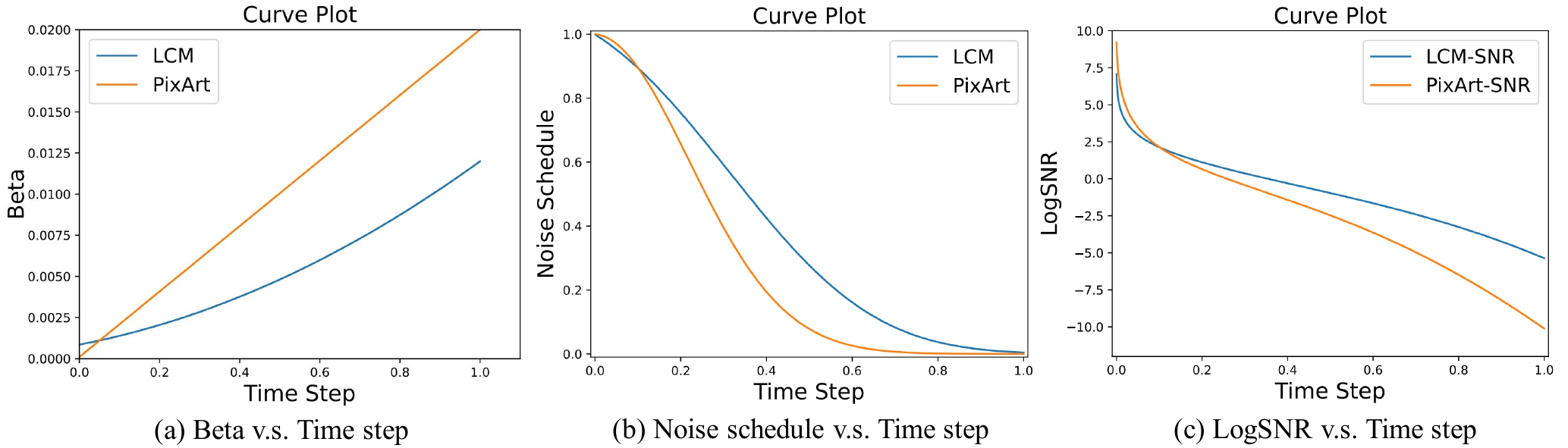}
\caption{Instantiations of $\beta_t$, noise schedule function and the corresponding logSNR between \model{} and LCM. $\beta_t$ is the coefficient in the diffusion process $z_t = \sqrt{\Bar{\alpha_t}}z_0 + \sqrt{1-\Bar{\alpha_t}}\epsilon, \alpha_t = 1 - \beta_t$.}
\label{fig:noise_snr}
\end{figure}

\noindent \textbf{Noise Schedule Adjustment.} Noise schedule is one of the most important parts of the diffusion process. Following~\citep{hoogeboom2023simple, chen2023importance}, we adapt the noise schedule function in LCM to align with the \pixarta{} noise schedule, which features a higher logSNR (signal-to-noise ratio) during the distillation training. Fig.~\ref{fig:noise_snr} visualizes the noise schedule functions under different choices of \model{} or LCM, along with their respective logSNR. Notably, \model{} can parameterize a broader range of noise distributions, a feature that has been shown further to enhance image generation~\citep{hoogeboom2023simple, chen2023importance}.

\subsection{Training efficiency and inference speedup \label{sec:train_and_inference}}
For training, as illustrated in Tab.~\ref{tab:pixart_lcm_train_compare}, we successfully conduct the distillation process within a 32GB GPU memory constraint, all while retaining the same batch size and supporting image resolution up to 1024 $\times$ 1024 with SDXL-LCM. Such training efficiency remarkably enables \model{} to be trained on a wide array of consumer-grade GPU specifications. In light of the discussions in Sec.\ref{sec:algorithm_and_modification}, regarding the beneficial impact of larger batch size, our method notably makes it feasible to utilize larger batch size even on GPUs with limited memory capacity.

\begin{figure}[!ht]
\centering
\footnotesize
\includegraphics[width=0.95\linewidth]{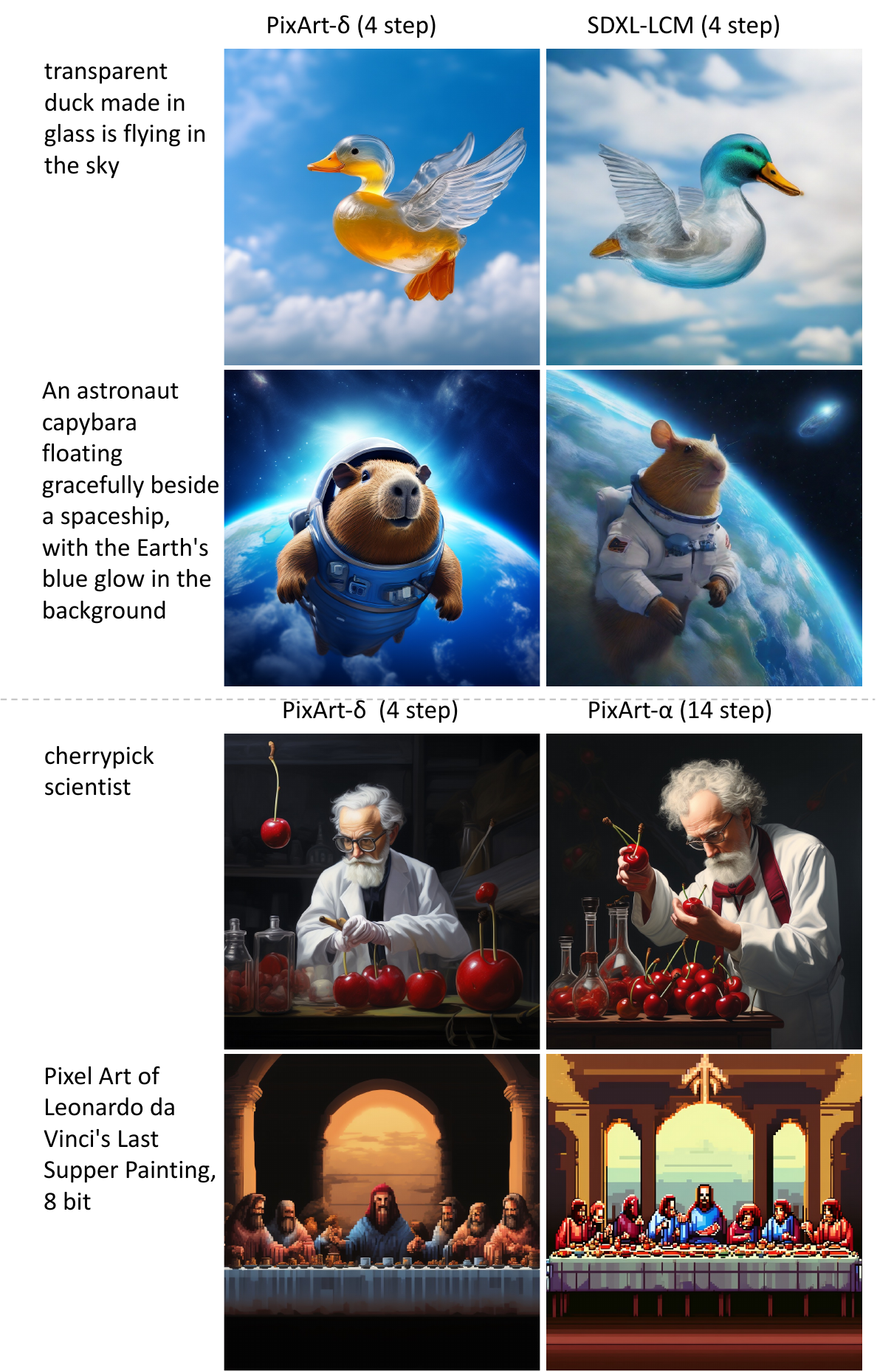}
\caption{
Examples of generated outputs. In the top half, the comparison is between \model{} and SDXL-LCM, with 4 sampling steps. In the bottom half, the comparison involves \model{} and \pixarta{} (teacher model, using DPM-Solver with 14 steps).
}
\label{fig:lcm_sample_compare}
\end{figure}

For inference, as shown in Tab.~\ref{tab:pixart_lcm_speed_compare} and Fig.~\ref{fig:lcm_sample_compare}, we present a comparative analysis of the generation speed achieved by our model, \model{}, against other methods like SDXL LCM-LoRA, \pixarta{}, and the SDXL standard across different hardware platforms. Consistently, \model{} achieves \textbf{1024x1024 high resolution} image generation within \textbf{0.5 seconds} on an A100, and also completes the process in a mere 3.3 seconds on a T4, 0.8 seconds on a V100, all with a batch size of 1. This is a significant improvement over the other methods, where, for instance, the SDXL standard takes up to 26.5 seconds on a T4 and 3.8 seconds on an A100. The efficiency of \model{} is evident as it maintains a consistent lead in generation speed with only 4 steps, compared to the 14 and 25 steps required by \pixarta{} and SDXL standard, respectively. Notably, with the implementation of 8-bit inference technology, \model{} requires less than \textbf{8GB of GPU VRAM}. This remarkable efficiency enables \model{} to operate on a wide range of GPU cards, and it even opens up the possibility of running on a CPU.

\begin{table}[ht]
% \begin{wraptable}{r}{8cm}
\centering
    \caption{Illustration of the training setting between LCM on \model{} and Stable Diffusion models. (* stands for Stable Diffusion Dreamshaper-v7 finetuned version)} 
\label{tab:pixart_lcm_train_compare}
\resizebox{0.7\linewidth}{!}{ 
\begin{tabular}{cccccc}
\toprule
% \multicolumn{1}{c}
Methods & \bf \model{} & SDXL LCM-LoRA  & SD-V1.5-LCM* \\
\midrule
Data Volume  & 120K & 650K & 650K \\
Resolution  & 1024px & 1024px & 768px \\
Batch Size & $12\times32$ & $12\times64$ & $16\times8$ \\
GPU Memory & $\sim$32G & $\sim$80G & $\sim$80G \\
\bottomrule

\end{tabular}
}
\vspace{-1em}
\end{table}
% \end{wraptable}

\begin{table}[ht]
% \begin{wraptable}{r}{8cm}
\centering
    \caption{Illustration of the generation speed we achieve on various devices. These tests are conducted on $1024\times1024$ resolution with a batch size of 1 in all cases. Corresponding image samples are shown in the Fig.~\ref{fig:lcm_sample_compare}} 
\label{tab:pixart_lcm_speed_compare}
\resizebox{0.8\linewidth}{!}{ 
\begin{tabular}{ccccccc}
\toprule
% \multicolumn{1}{c}
\multirow{2}{*}{\bf Hardware} & \bf \model{} & SDXL LCM-LoRA  & \pixarta{} & SDXL standard \\
\cmidrule(lr){2-5}
& 4 steps & 4 steps & 14 steps & 25 steps \\
\midrule
T4  & 3.3s & 8.4s & 16.0s & 26.5s \\
V100 & 0.8s & 1.2s & 5.5s & 7.7s \\
A100 & 0.5s & 1.2s & 2.2s & 3.8s \\
\bottomrule

\end{tabular}
}
\vspace{-1em}
\end{table}
% \end{wraptable}

\subsection{Training Details \label{sec:training_details}} 
As discussed in Sec.~\ref{sec:algorithm_and_modification}, we conduct our experiments in two resolution settings, 512×512 and 1024×1024, utilizing a high-quality internal dataset with 120K images. We smoothly train the models in both resolutions by leveraging the multi-scale image generation capabilities of \pixarta{}, which supports 512px and 1024px resolutions. For both resolutions, \model{} yields impressive results before reaching 5K iterations, with only minimal improvements observed thereafter. The training is executed on 2 V100 GPUs with a total batch size of 24, a learning rate of 2e-5, EMA rate $\mu = 0.95$, and using AdamW optimizer~\citep{loshchilov2017adamw}. We employ DDIM-Solver~\citep{song2023consistency} and a skipping step $k=20$~\citep{luo2023lcm} for efficiency.
As noted in Sec.~\ref{sec:algorithm_and_modification} and illustrated in Fig.~\ref{fig:noise_snr}, modifications are made to the original LCM scheduler to accommodate differences between the pre-trained \pixarta{} and Stable Diffusion models. Following the \pixarta{} approach, we alter the $\beta_t$ in the diffusion process from a scaled linear to a linear curve, adjusting $\beta_{t_0}$ from 0.00085 to 0.0001, and $\beta_{t_T}$ from 0.012 and to 0.02 at the same time. 
The guidance scale $\omega_{fix}$ is set to 4.5, identified as optimal in \pixarta{}. While omitting the Fourier embedding of $\omega$ in LCM during training, both \pixarta{} and \model{} maintain identical structures and trainable parameters. This allows us to initialize the consistency function $\vf_{\vtheta}(\hat{\vz},\omega_{fix},\vc, t_n)$ with the same parameters as the teacher diffusion model (\pixarta{}) without compromising performance. Building on the success of LCM-LoRA~\citep{luo2023lcm}, \model{} can further easily integrate LCM-LoRA, enhancing its adaptability for a more diverse range of applications.

\begin{figure}[ht]
\begin{center}
\footnotesize
\includegraphics[width=1\linewidth]{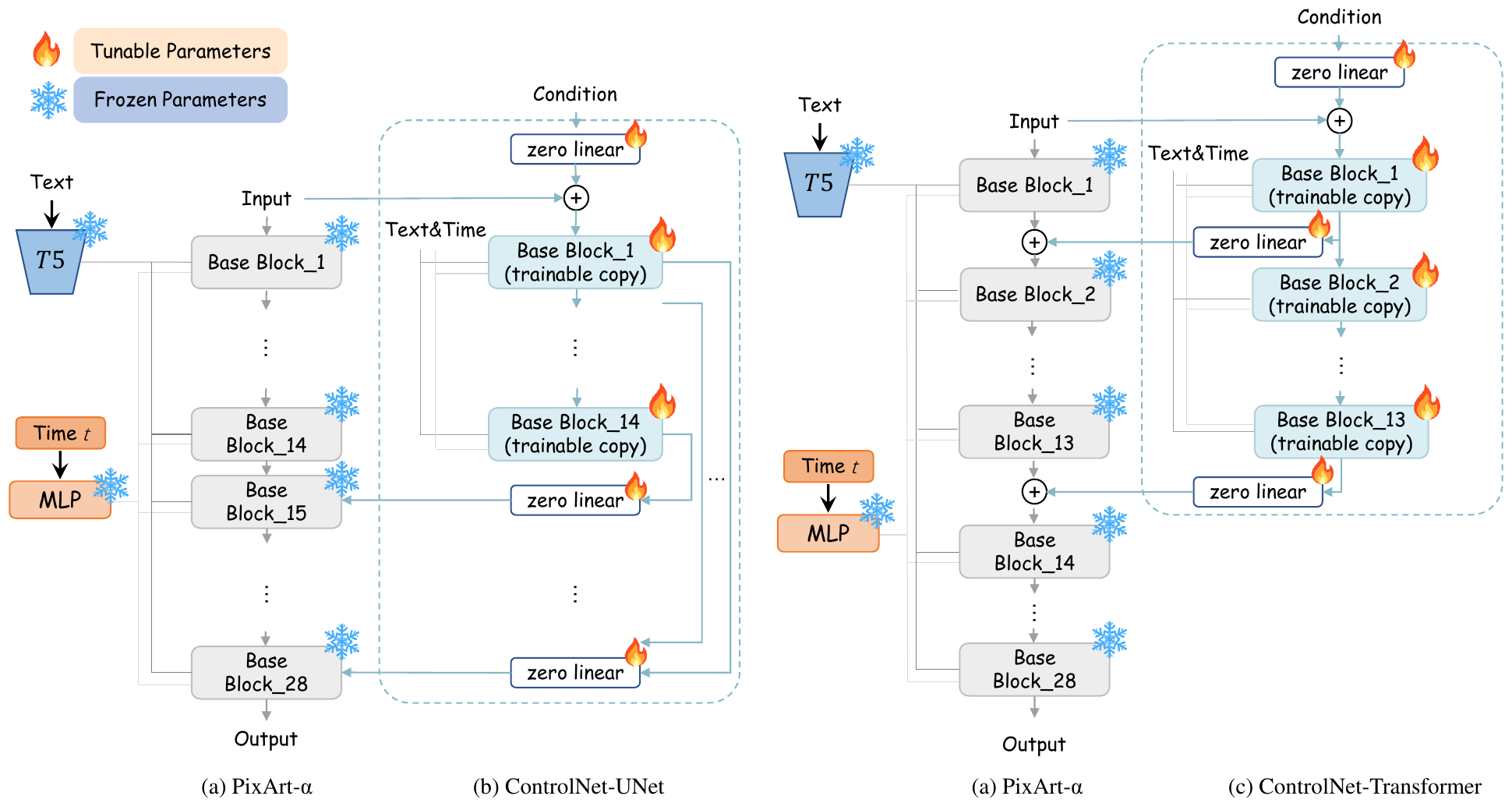}
\caption{\model integrated with ControlNet. (b): ControlNet-UNet. Base blocks are categorized into ``encoder'' and ``decoder'' stages. The controlnet structure is applied to each encoder level of \model, and the output is connected to the decoder stage via skip-connections. (c): ControlNet-Transformer. The ControlNet is applied to the first several blocks. The output of each block is added to the output of the corresponding frozen block, serving as the input of the next frozen block.} 
\label{fig:controlnet-architecture}
\end{center} 
\end{figure}

\section{ControlNet in \model}
\subsection{Architecture}
ControlNet, initially tailored for the UNet architecture, employed skip connections to enhance the integration of control signals. 
The seamless incorporation of ControlNet into Transformer-based models, exemplified by \model, introduces a distinctive challenge. Unlike UNet, Transformers lack explicit ``encoder" and ``decoder" blocks, making the conventional connection between these components inappropriate.

In response to this challenge, we propose an innovative approach, ControlNet-Transformer, to ensure the effective integration of ControlNet with Transformers, preserving ControlNet's effectiveness in managing control information and high-quality generation of \model.

\model contains 28 Transformer blocks. We replace the original zero-convolution in ControlNet with a zero linear layer, that is, a linear layer with both weight and bias initialized to zero. We explore the following network architectures:

\begin{itemize}
    \item \textbf{ControlNet-UNet}~\citep{zhang2023controlnet}. To follow the original ControlNet design, we treat the first 14 blocks as the ``encoder'' level of \model, and the last 14 blocks as the ``decoder'' level of \model. We use ControlNet to create a trainable copy of the 14 encoding blocks. Subsequently, the outputs from these blocks are integrated by addition into the 14 skip-connections, which link to the last 14 decoder blocks. The network design is shown in Fig.~\ref{fig:controlnet-architecture} (b). 

    It is crucial to note that this adaptation, referred to as ControlNet-UNet, encounters challenges due to the absence of explicit ``encoder" and ``decoder" stages and skip-connections in the original Transformer design. This adaptation departs from the conventional architecture of the Transformer, which hampers the effectiveness and results in suboptimal outcomes.
    
    \item \textbf{ControlNet-Transformer}. To address these challenges, we propose a novel and specifically tailored design for Transformers, illustrated in Fig.~\ref{fig:controlnet-architecture} (c). This innovative approach aims to seamlessly integrate the ControlNet structure with the inherent characteristics of Transformer architectures. To achieve this integration, we selectively apply the ControlNet structure to the initial $N$ base blocks. In this context, we generate $N$ trainable copies of the first $N$ base blocks. The output of $i^{th}$ trainable block is intricately connected to a zero linear layer, and the resulting output is then added to the output of the corresponding $i^{th}$ frozen block. Subsequently, this combined output serves as the input for the subsequent $(i+1)^{th}$ frozen block. This design adheres to the original data flow of PixArt, and our observations underscore the significant enhancement in controllability and performance achieved by ControlNet-Transformer. This approach represents a crucial step toward harnessing the full potential of Transformer-based models in such applications. The ablation study of $N$ is described in Sec.~\ref{controlnet:ablation}, and we use $N=13$ as the final model.
\end{itemize}

\subsection{Experiment Settings}
We use a HED edge map in \model as the condition and conduct an ablation study on 512px generation, focusing on network architecture variations. Specifically, we conduct ablations on both the ControlNet-UNet and ControlNet-Transformer. Other conditions, such as canny, will be a future work.
For ControlNet-Transformer, we ablate the number of copied blocks, including 1, 4, 7, 13, and 27. We extract the HED on the internal data, and the gradient accumulation step is set as 4 following ~\citep{zhang2023controlnet}'s advice that recommendation that larger gradient accumulation leads to improved results. The optimizer and learning rate are set as the same setting of \model. All the experiments are conducted on 16 V100 GPUs with 32GB. The batch size per GPU for experiment ControlNet-Transformer ($N=27$) is set as 2. For all other experiments, the batch size is set as 12. Our training set consists of 3M HED and image pairs.

\begin{figure}[ht]
\begin{center}
\footnotesize
\includegraphics[width=1\linewidth]{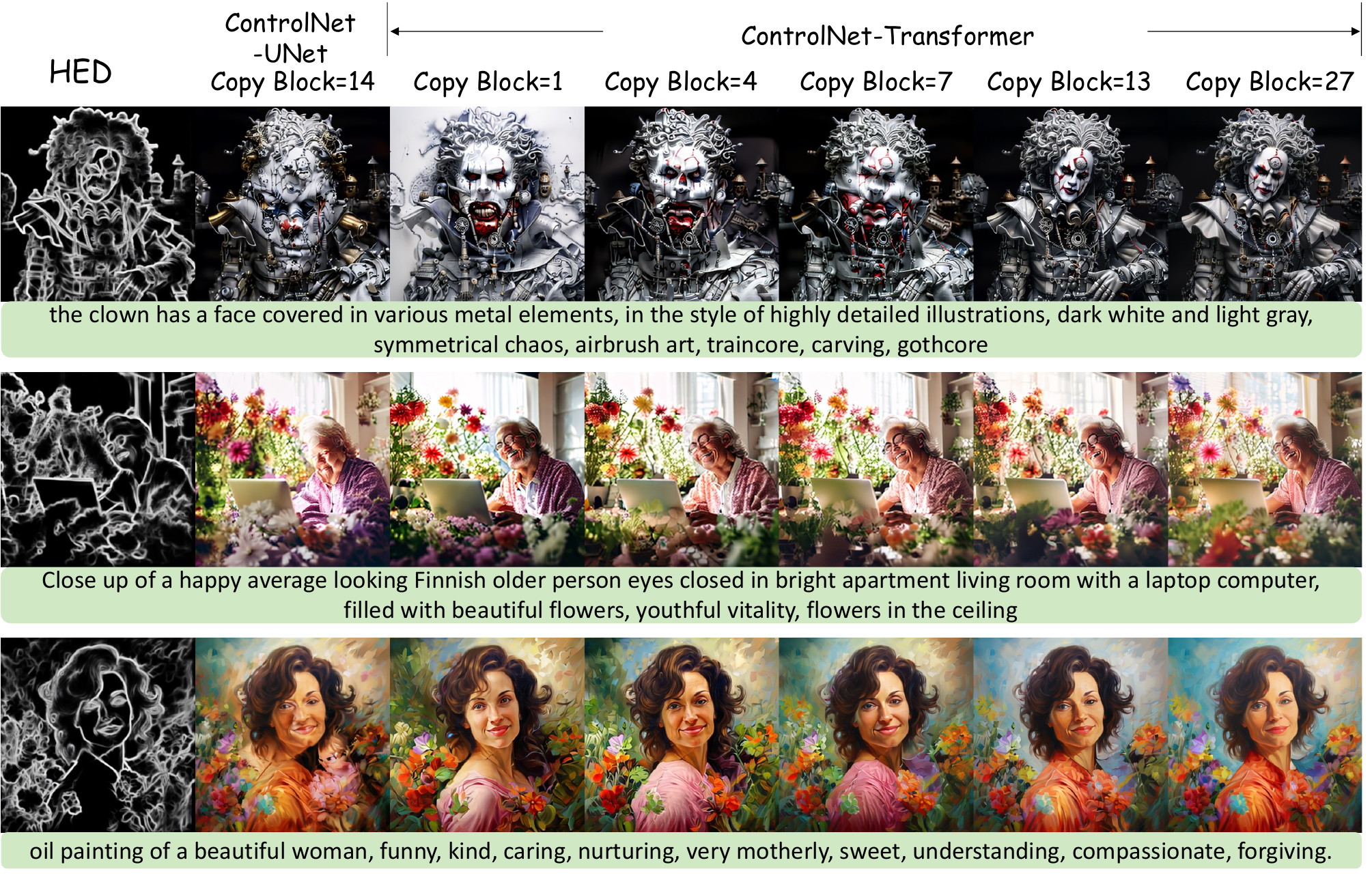}
\caption{The ablation study of ControlNet-UNet and ControlNet-Transformer. ControlNet-Transformer yields much better results than ControlNet-UNet. The controllability of ControlNet-Transformer increases as the number of copy blocks increases.
} 
\label{fig:controlnet-copy}
\end{center} 
\end{figure}

\subsection{Ablation Study}\label{controlnet:ablation}
As shown in Fig.~\ref{fig:controlnet-copy}, ControlNet-Transformer generally outperforms, demonstrating faster convergence and improved overall performance. This superiority can be attributed to the fact that ControlNet-Transformer's design aligns seamlessly with the inherent data flow of Transformer architectures. Conversely, ControlNet-UNet introduces a conceptual information flow between the non-existing ``encoder" and ``decoder" stages, deviating from the Transformer's natural data processing pattern. 

In our ablation study concerning the number of copied blocks, we observe that for the majority of scenarios, such as scenes and objects, satisfactory results can be achieved with merely $N=1$.
However, in challenging edge conditions, such as the outline edge of human faces and bodies, performance tends to improve as $N$ increases.
Considering a balance between computational burden and performance, we find that $N=13$ is the optimal choice in our final design.

\subsection{Convergence}
\begin{figure}[ht]
\begin{center}
\footnotesize
\includegraphics[width=1\linewidth]{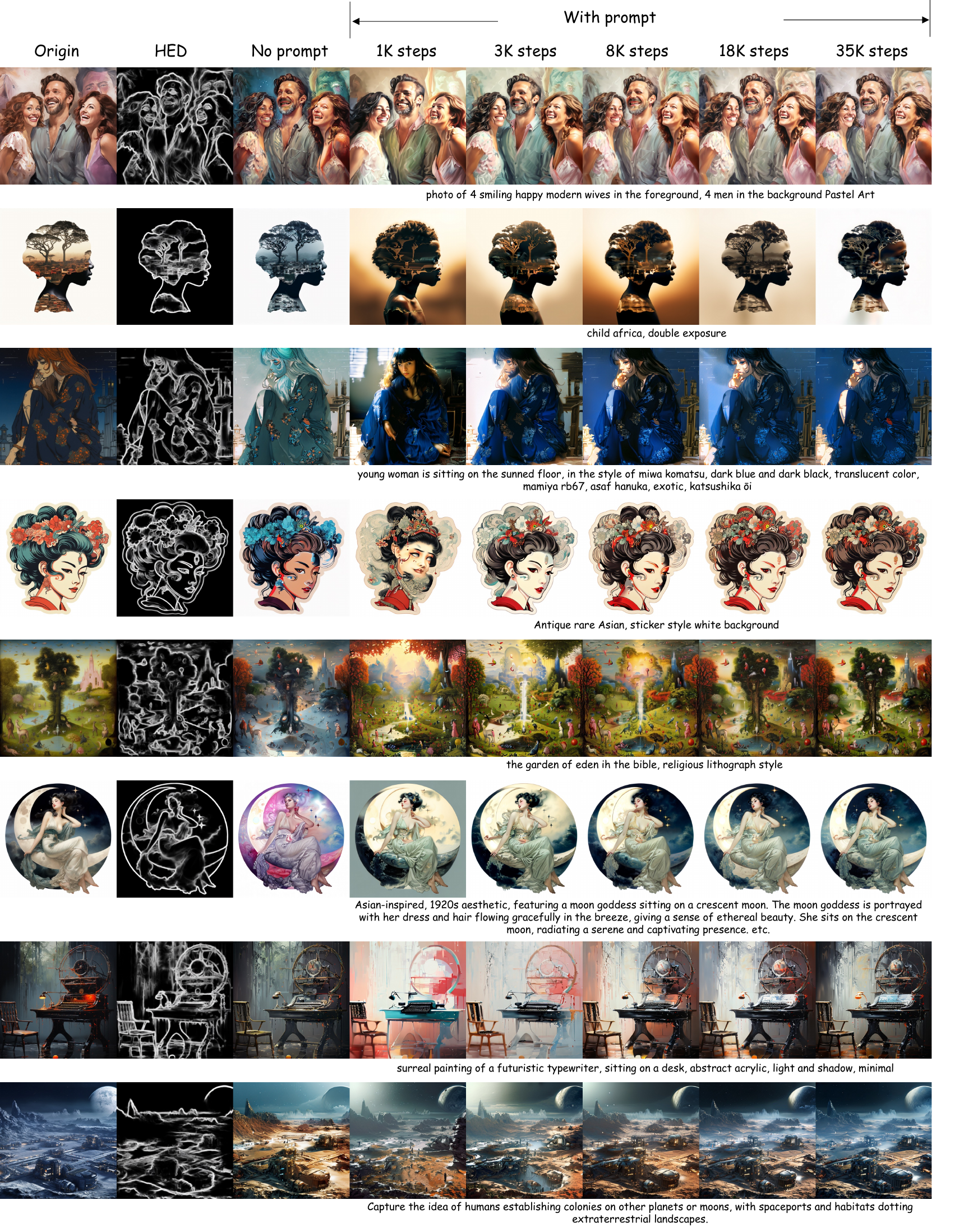}
\caption{The influence of training steps. The convergence is fast, with details progressively improving and aligning more closely with the HED edge map as the training steps increase.} 
\label{fig:controlnet-convergence}
\end{center} 
\end{figure}

As described in Fig.~\ref{fig:controlnet-convergence}, we analyze the effect of training steps. The experiment is conducted on ControlNet-Transformer ($N=13$). From our observation, the convergence is very fast, with most edges achieving satisfactory results at around 1,000 training steps. Moreover, we note a gradual improvement in results as the number of training steps increases, particularly noticeable in enhancing the quality of outline edges for human faces and bodies.
This observation underscores the efficiency and effectiveness of ControlNet-Transformer.

We observe a similar ``sudden converge'' phenomenon in our model, as also observed in the original ControlNet work, where it ``suddenly'' adapts to the training conditions. Empirical observations indicate that this phenomenon typically occurs between 300 to 1,000 steps, with the convergence steps being influenced by the difficulty level of the specified conditions. Simpler edges tend to converge at earlier steps, while more challenging edges require additional steps for convergence. After ``sudden converge'', we observe an improvement in details as the number of steps increases.

\begin{figure}[htb]
\begin{center}
\footnotesize
\includegraphics[width=1\linewidth]{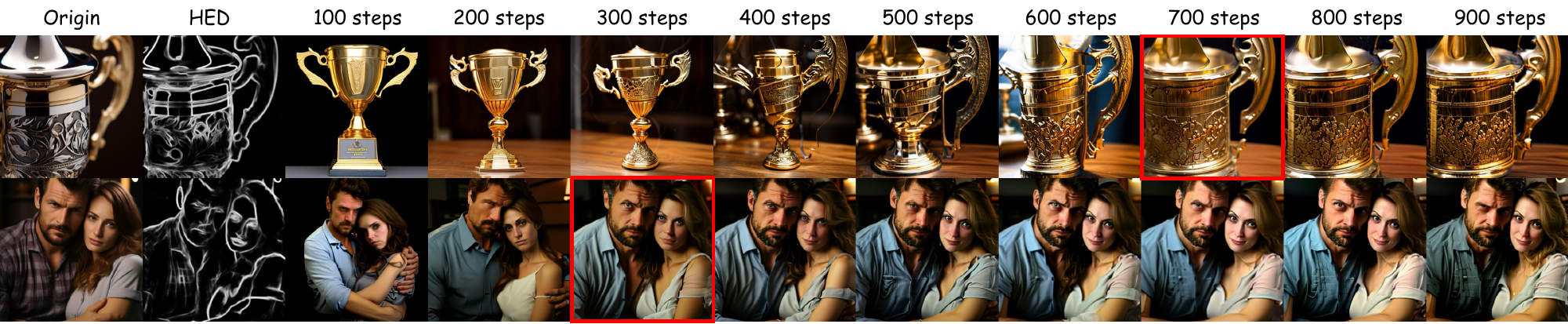}
\caption{Example of ``Sudden  Converge'' during PixArt-ControlNet training. We empirically observe it happens before 1000 iterations.} 
\label{fig:sudden-convergence}
\end{center} 
\end{figure}

\begin{figure}[ht]
\begin{center}
\footnotesize
\includegraphics[width=1\linewidth]{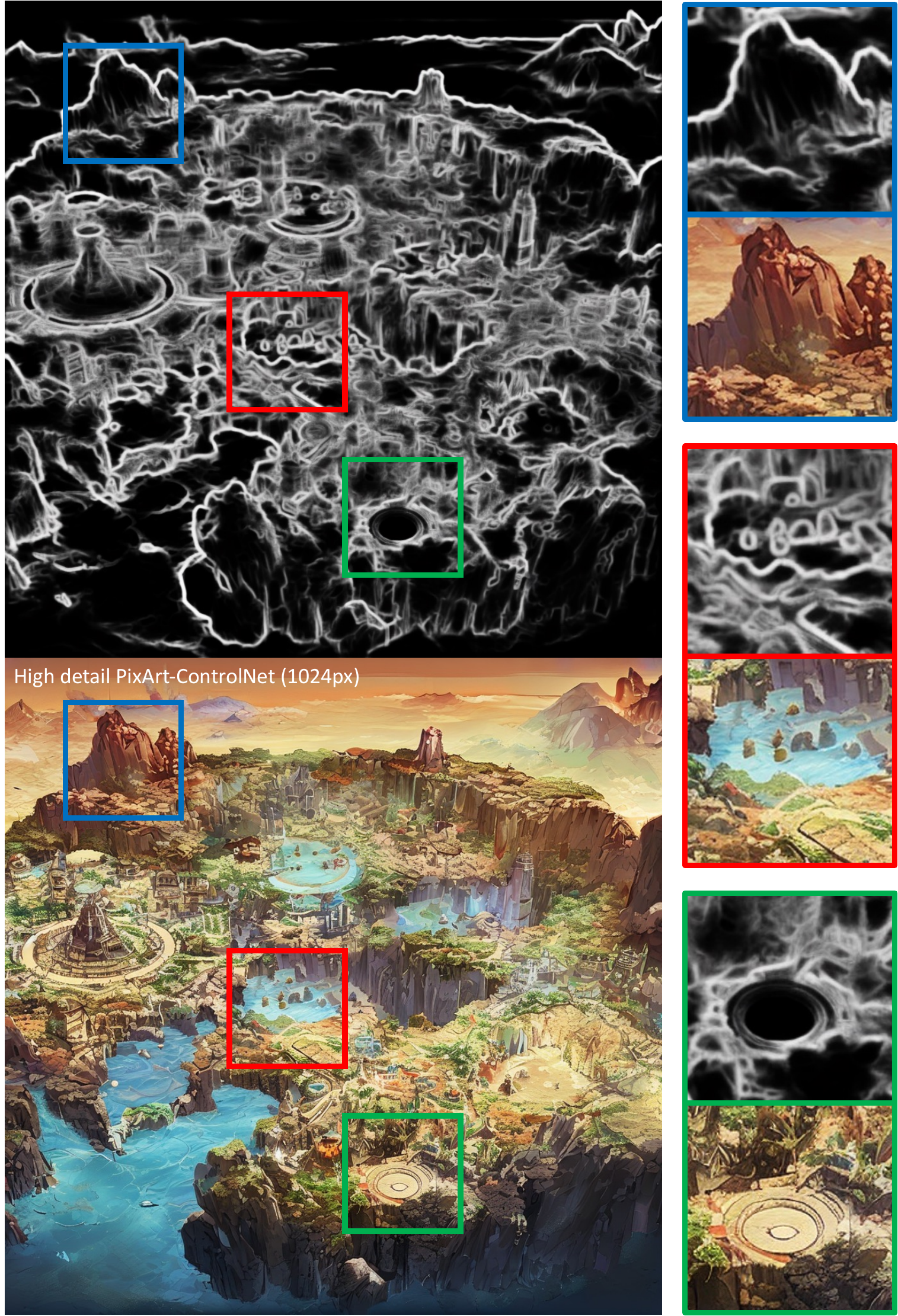}
\caption{
High-resolution and fine-grained controllable image generation.
The output is generated with the prompt ``the map of the final fantasy game's main island, in the style of hirohiko araki, raymond swanland, monumental murals, mosaics, naturalistic rendering, vorticism, use of earth tones.''} 
\label{fig:zoom1}
\end{center} 
\end{figure}

\begin{figure}[ht]
\begin{center}
\footnotesize
\includegraphics[width=1\linewidth]{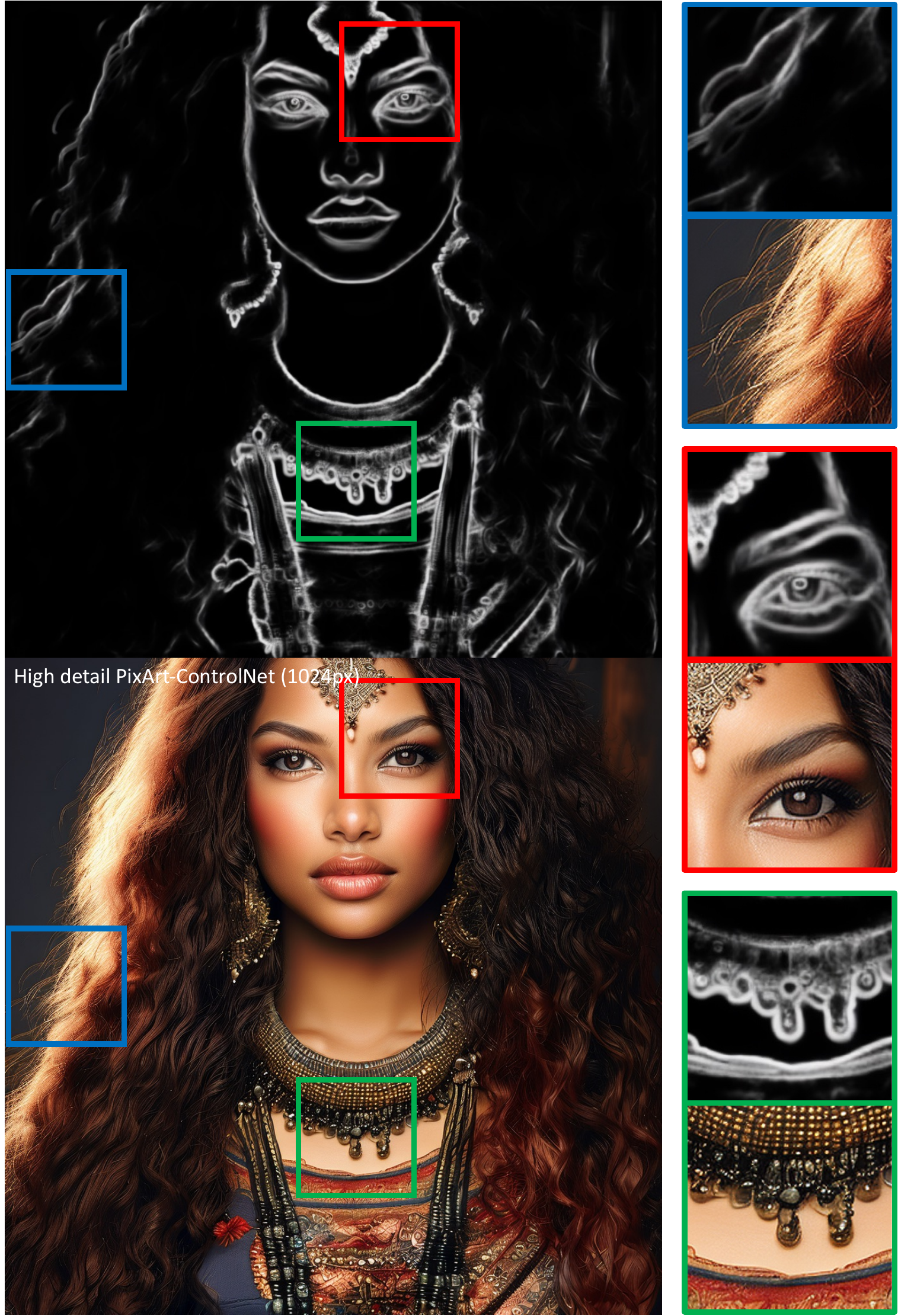}
\caption{High-resolution and fine-grained controllable image generation.
The output is generated with the prompt ``Multicultural beauty. Women of different ethnicity - Caucasian, African, Asian and Indian.''} 
\label{fig:zoom2}
\end{center} 
\end{figure}

\begin{figure}[ht]
\begin{center}
\footnotesize
\includegraphics[width=1\linewidth]{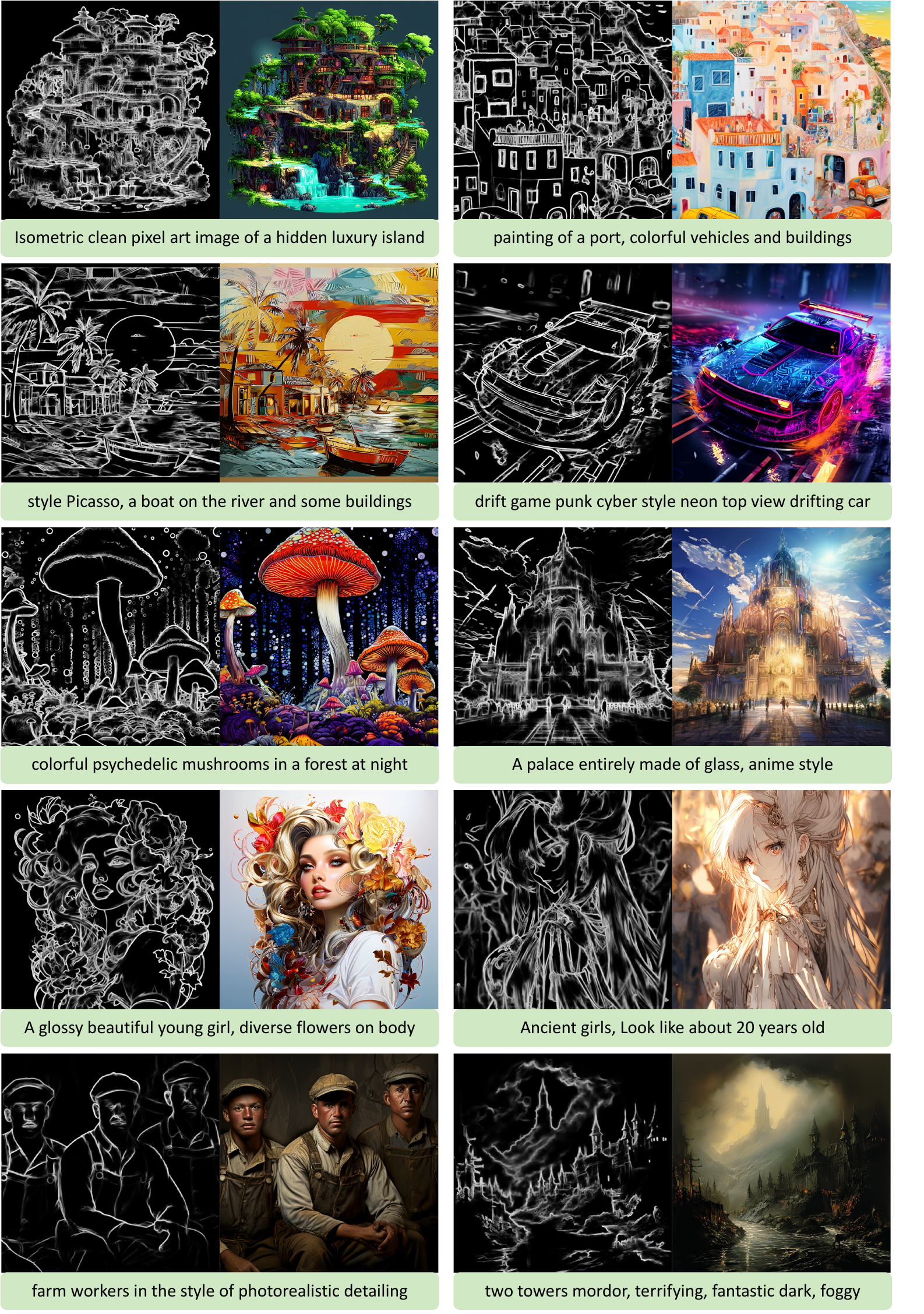}
\caption{More examples of our PixArt-ControlNet generated images.} 
\label{fig:1024px_diverse}
\end{center} 
\end{figure}

\subsection{1024px Results}
Building upon the powerful text-to-image generation framework of PixArt, our proposed PixArt-ControlNet extends these capabilities to produce high-resolution images with a granular level of control. This is vividly demonstrated in the detailed visualizations presented in Fig.~\ref{fig:zoom1} and Fig.~\ref{fig:zoom2}. Upon closer inspection of these figures, it is apparent that PixArt-ControlNet can exert precise control over the geometric composition of the resultant images, achieving fidelity down to individual strands of hair.

\section{Conclusion}

In this report, we present \model, a better text-to-image generation model integrating Latent Consistency Models (LCM) to achieve 4-step sampling acceleration while maintaining high quality. 
We also propose Transformer-based ControlNet, a specialized design tailored for Transformer architecture, enabling precise control over generated images. 
Through extensive experiments, we demonstrate \model's faster sampling and ControlNet-Transformer's effectiveness in high-resolution and controlled image generation. 
Our model can generate high-quality 1024px and fine-grained controllable images in 1 second.
\model pushes the state-of-the-art in faster and more controlled image generation, unlocking new capabilities for real-time applications. 

\paragraph{Acknowledgement.} We extend our sincere gratitude to Patrick von Platen and Suraj Patil from Hugging Face for their invaluable support and contributions to this work.
\newpage

\clearpage

\clearpage
\bibliography{iclr2024_conference}

\begin{thebibliography}{13}
\providecommand{\natexlab}[1]{#1}
\providecommand{\url}[1]{\texttt{#1}}
\expandafter\ifx\csname urlstyle\endcsname\relax
  \providecommand{\doi}[1]{doi: #1}\else
  \providecommand{\doi}{doi: \begingroup \urlstyle{rm}\Url}\fi

\bibitem[Chen et~al.(2023)Chen, Yu, Ge, Yao, Xie, Wu, Wang, Kwok, Luo, Lu, et~al.]{chen2023pixart}
Junsong Chen, Jincheng Yu, Chongjian Ge, Lewei Yao, Enze Xie, Yue Wu, Zhongdao Wang, James Kwok, Ping Luo, Huchuan Lu, et~al.
\newblock Pixart-$\alpha$: Fast training of diffusion transformer for photorealistic text-to-image synthesis.
\newblock \emph{arXiv preprint arXiv:2310.00426}, 2023.

\bibitem[Chen(2023)]{chen2023importance}
Ting Chen.
\newblock On the importance of noise scheduling for diffusion models.
\newblock \emph{arXiv preprint arXiv:2301.10972}, 2023.

\bibitem[Deng et~al.(2009)Deng, Dong, Socher, Li, Li, and Fei-Fei]{deng2009imagenet}
Jia Deng, Wei Dong, Richard Socher, Li-Jia Li, Kai Li, and Li~Fei-Fei.
\newblock Imagenet: A large-scale hierarchical image database.
\newblock In \emph{2009 IEEE conference on computer vision and pattern recognition}, pp.\  248--255. Ieee, 2009.

\bibitem[Hoogeboom et~al.(2023)Hoogeboom, Heek, and Salimans]{hoogeboom2023simple}
Emiel Hoogeboom, Jonathan Heek, and Tim Salimans.
\newblock simple diffusion: End-to-end diffusion for high resolution images.
\newblock \emph{arXiv preprint arXiv:2301.11093}, 2023.

\bibitem[Hu et~al.(2021)Hu, Wallis, Allen-Zhu, Li, Wang, Wang, Chen, et~al.]{hu2021lora}
Edward~J Hu, Phillip Wallis, Zeyuan Allen-Zhu, Yuanzhi Li, Shean Wang, Lu~Wang, Weizhu Chen, et~al.
\newblock Lora: Low-rank adaptation of large language models.
\newblock In \emph{ICLR}, 2021.

\bibitem[Loshchilov \& Hutter(2017)Loshchilov and Hutter]{loshchilov2017adamw}
Ilya Loshchilov and Frank Hutter.
\newblock Decoupled weight decay regularization.
\newblock In \emph{arXiv}, 2017.

\bibitem[Luo et~al.(2023{\natexlab{a}})Luo, Tan, Huang, Li, and Zhao]{luo2023latent}
Simian Luo, Yiqin Tan, Longbo Huang, Jian Li, and Hang Zhao.
\newblock Latent consistency models: Synthesizing high-resolution images with few-step inference.
\newblock \emph{arXiv preprint arXiv:2310.04378}, 2023{\natexlab{a}}.

\bibitem[Luo et~al.(2023{\natexlab{b}})Luo, Tan, Patil, Gu, von Platen, Passos, Huang, Li, and Zhao]{luo2023lcm}
Simian Luo, Yiqin Tan, Suraj Patil, Daniel Gu, Patrick von Platen, Apolin{\'a}rio Passos, Longbo Huang, Jian Li, and Hang Zhao.
\newblock Lcm-lora: A universal stable-diffusion acceleration module.
\newblock \emph{arXiv preprint arXiv:2311.05556}, 2023{\natexlab{b}}.

\bibitem[Podell et~al.(2023)Podell, English, Lacey, Blattmann, Dockhorn, Müller, Penna, and Rombach]{podell2023sdxl}
Dustin Podell, Zion English, Kyle Lacey, Andreas Blattmann, Tim Dockhorn, Jonas Müller, Joe Penna, and Robin Rombach.
\newblock Sdxl: Improving latent diffusion models for high-resolution image synthesis.
\newblock In \emph{arXiv}, 2023.

\bibitem[Rombach et~al.(2022)Rombach, Blattmann, Lorenz, Esser, and Ommer]{rombach2022high}
Robin Rombach, Andreas Blattmann, Dominik Lorenz, Patrick Esser, and Bj{\"o}rn Ommer.
\newblock High-resolution image synthesis with latent diffusion models.
\newblock In \emph{CVPR}, 2022.

\bibitem[Song et~al.(2023)Song, Dhariwal, Chen, and Sutskever]{song2023consistency}
Yang Song, Prafulla Dhariwal, Mark Chen, and Ilya Sutskever.
\newblock Consistency models.
\newblock \emph{arXiv preprint arXiv:2303.01469}, 2023.

\bibitem[von Platen et~al.(2023)von Platen, Patil, Lozhkov, Cuenca, Lambert, Rasul, Davaadorj, and Wolf]{diffusers}
Patrick von Platen, Suraj Patil, Anton Lozhkov, Pedro Cuenca, Nathan Lambert, Kashif Rasul, Mishig Davaadorj, and Thomas Wolf.
\newblock Diffusers: State-of-the-art diffusion models, 2023.
\newblock URL \url{https://huggingface.co/docs/diffusers/main/en/api/pipelines/pixart#inference-with-under-8gb-gpu-vram?}

\bibitem[Zhang et~al.(2023)Zhang, Rao, and Agrawala]{zhang2023controlnet}
Lvmin Zhang, Anyi Rao, and Maneesh Agrawala.
\newblock Adding conditional control to text-to-image diffusion models, 2023.

\end{thebibliography}
\bibliographystyle{iclr2024_conference}

\end{document}